\documentclass[runningheads, twoside]{llncs}
\usepackage{caption}
\usepackage{subcaption}
\usepackage{graphicx}                   
\usepackage{multicol}                   
\usepackage{multirow}                   
\usepackage{rotating}                   
\usepackage{tabularx}                   
\usepackage{amsmath}
\usepackage{amssymb}
\usepackage{physics}
\usepackage{hyperref}
\hypersetup{
     colorlinks=true,
     linkcolor=blue,
     filecolor=magenta,
     citecolor = black,      
     urlcolor=cyan,
     }

\begin{document}
\title{Model Predictive Control with Self-supervised Representation Learning}
\titlerunning{MPC with SRL}
%
%
\author{Jonas Matthies \and
Muhammad Burhan Hafez \and
Mostafa Kotb \and Stefan Wermter}
\authorrunning{J. Matthies et al.}
%
\institute{Knowledge Technology, Department of Informatics, University of Hamburg \email{Jonas-Matthies@gmx.de, \{burhan.hafez,mostafa.kotb,stefan.wermter\}@uni-hamburg.de}}

\maketitle              
\begin{abstract}
Over the last few years, we have not seen any major developments in model-free or model-based learning methods that would make one obsolete relative to the other. In most cases, the used technique is heavily dependent on the use case scenario or other attributes, e.g. the environment. Both approaches have their own advantages, for example, sample efficiency or computational efficiency. However, when combining the two, the advantages of each can be combined and hence achieve better performance. The TD-MPC framework is an example of this approach. On the one hand, a world model in combination with model predictive control is used to get a good initial estimate of the value function. On the other hand, a Q function is used to provide a good long-term estimate. Similar to algorithms like MuZero a latent state representation is used, where only task-relevant information is encoded to reduce the complexity. In this paper, we propose the use of a reconstruction function within the TD-MPC framework, so that the agent can reconstruct the original observation given the internal state representation. This allows our agent to have a more stable learning signal during training and also improves sample efficiency. Our proposed addition of another loss term leads to improved performance on both state- and image-based tasks from the DeepMind-Control suite. 

\end{abstract}


\section{Introduction}
In the domain of Reinforcement Learning (RL), there are two main categories of approaches, namely the \textit{model-free} and \textit{model-based} learning methods. Model-free methods are concerned with learning the quality of specific states or pairs of state and action, whereas model-based methods learn a model of the environments dynamics. While model-free approaches evaluate the possible actions and next states at each time step, model-based methods use a process called planning. With the help of the world model, the actor can simulate a sequence of actions by mimicking the real environment and hence create and evaluate different plans. However, planning is often restrained to short horizons because extensive planning is not feasible due to the model becoming less and less accurate the more actions we simulate.

Inherently, both model-free and model-based approaches have their own strengths and weaknesses, prompting the question of whether the two methods can be combined to create one framework that leverages the benefits of both. Although research in this area is still in its early stages, there have already been encouraging findings indicating an increased performance for these hybrid learning approaches \cite{DBLP:journals/corr/abs-1911-08265,NEURIPS2021_d5eca8dc,pmlr-v164-sikchi22a,Hafner2020Dream,8852254,HAFEZ2020103630}. The recently proposed Temporal Difference Learning for Model Predictive Control (TD-MPC)\cite{Hansen2022tdmpc} framework, for example, demonstrated that the combination of model-free and model-based methods can combine the advantages of both methods and is able to achieve superior sample efficiency as well as improved performance when compared to other state-of-the-art algorithms such as Soft Actor-Critic (SAC)\cite{pmlr-v80-haarnoja18b} or LOOP \cite{pmlr-v164-sikchi22a} on a variety of tasks. It utilizes a combination of Model Predictive Control (MPC) for short-horizon planning and  a Q-value function for long-term reward estimates. Similar to other state-of-the-art algorithms like MuZero \cite{DBLP:journals/corr/abs-1911-08265}, Dreamer \cite{Hafner2020Dream} or EfficientZero \cite{NEURIPS2021_d5eca8dc}, latent representations of the environment states are extensively used during training to provide the agent with an abstract representation and thus reduce the complexity and focus on task-relevant details.

In this paper, we propose the use of an additional loss term for the training of the Task-Oriented Latent Dynamics (TOLD)\cite{Hansen2022tdmpc} model, which supplements the already implemented consistency loss. The additional term is calculated with the help of the reconstruction function, which aims to reconstruct the original observation from the latent space representation. While the latent space is designed to omit irrelevant details from the observations, the reconstruction function on the other hand is used to restore the original observation. The use of a reconstruction function has already proven to be performance-enhancing in various other algorithms like PlaNet \cite{pmlr-v97-hafner19a}, Dreamer \cite{Hafner2020Dream}, Dreamer-v2 \cite{hafner2021mastering} or a modified version of MuZero \cite{DBLP:journals/corr/abs-2102-05599}. Similar to the rest of the TOLD model, the parameters of the reconstruction function are learnt jointly with the representation, dynamics, reward and value functions using Temporal Difference (TD) learning. By complementing the consistency loss with a reconstruction loss, the aim is to stabilize the learning process of the agent through an enhanced learning signal. We seek to answer the question of how our version of TD-MPC with Self-supervised Representation Learning (SRL) compares to the original implementation of the framework. An implementation of the adopted TD-MPC framework can be found at \url{https://github.com/Jonas-SM/TD-MPC-SRL/}.

\section{Background}

\subsection{Formal Definition}
We define an infinite-horizon Markov-Decission Process (MDP) characterized by the tuple $(S,A,T,R,\gamma,p_0)$ to model the agent-environment interaction. The set of possible states and actions in the environment are defined as $S$ and $A$ respectively, where $S \in \mathbb{R}^n$ and $A \in \mathbb{R}^m$ are continuous state and action spaces. The dynamics of the environment can be defined through the transition function $T : S \times A \times S \mapsto [0,1]$. It determines the probability of reaching state $s'$ when taking action $a$ in state $s$. Similarly, we define our reward function as $R : S \times A \mapsto \mathbb{R}$. Hence this function maps every pair of state and action to a scalar value which represents the immediate feedback an agent receives after taking an action. Additionally, $\gamma: \gamma \in [0,1)$ is a discount factor and $p_0$ denotes the initial state distribution of the environment, which specifies the probability distribution of the agent starting in each possible state. This tuple constitutes the complete MDP and serves as the formal framework for the TD-MPC algorithm. Another essential component of an RL agent is the policy $\pi : S \mapsto A$, which is a mapping from states to actions that is iteratively updated throughout training. At every timestep, this policy determines which action the agent executes and thus defines the agent's behaviour. Over the course of training, we aim to learn a parameterized mapping $\pi_\theta$, such that the agent achieves a maximum reward. 

\subsection{TOLD model}
The purpose of the TOLD model is to provide the agent with an abstract model of the world, that excludes all information irrelevant to the task \cite{Hansen2022tdmpc}. We will also often refer to this abstract representation of the world as the \textit{latent space}. Additionally, as a result of removing unnecessary details and, thus in most cases, significantly simplifying the environment's complexity, the training process will also become less complicated for the agent \cite{NEURIPS2018_2de5d166}. With the help of the models' different components, a number of quantities can be predicted that steer the learning process. The model itself consists of five components \cite{Hansen2022tdmpc}:

\begin{enumerate}
    \item \textbf{Representation function:} $z_t = h_\theta(s_t)$, simple encoder that encodes an observation $s_t$ into its respective latent state $z_t$.
    \item \textbf{Dynamics function:} $z_{t+1} = d_\theta(z_t, a_t)$, model of the world used to predict the subsequent latent state $z_{t+1}$.
	\item \textbf{Reward function:} $\hat{r}_t = R_{\theta}(z_t,a_t)$, predicts the expected reward $\hat{r}_t$ for a latent state $z_t$ and action $a_t$.
	\item \textbf{Value function:} $\hat{q}_t = Q_\theta(z_t,a_t)$, calculates an estimate of the expected return $\hat{q}_t$ given latent state $z_t$ and action $a_t$.
	\item \textbf{Policy function:} $\hat{a}_t = \pi_\theta(z_t)$, predicts the best action $\hat{a}_t$ to take for a given latent state $z_t$.
\end{enumerate}

All components are implemented using purely deterministic Multilayer Perceptrons (MLPs) and $\theta$ denotes the current parameterization of the TOLD model \cite{Hansen2022tdmpc}. Similar to other approaches like the MuZero algorithm \cite{DBLP:journals/corr/abs-1911-08265} that also utilizes both a model of the world and a value function, TD-MPC learns a policy $\pi_\theta$ additional to the model of the environment \cite{Hansen2022tdmpc}. This policy is  involved in the evaluation of the TOLD model, as well as the planning process of the agent to generate sample trajectories. In addition to the regular TOLD model, TD-MPC utilizes a target network, which is essentially a regular copy of the TOLD model, whose parameters $\theta^-$ are a slow-moving average of $\theta$ \cite{Hansen2022tdmpc}. The use of a target network can help with preventing the learning process from becoming unstable and is used in many Deep-Q RL algorithms such as Deep Q-learning \cite{DBLP:journals/corr/MnihKSGAWR13}, TD3 \cite{pmlr-v80-fujimoto18a} and DDPG \cite{journals/corr/LillicrapHPHETS15}. 
As it is a slow-moving average of $\theta$ it is also updated at the same time according to the following rule with $\zeta \in [0,1)$ being a constant \cite{Hansen2022tdmpc}:

\begin{equation}
\theta_{t+1}^- = (1 - \zeta) \theta_t^{-} + \zeta \theta_t. 
\end{equation}

During training, the TOLD model is updated by minimizing the loss 

\begin{equation}
\label{eq:totalloss}
\mathcal{J}(\theta) = \sum_{i=t}^{t+H} \lambda^{i-t}\mathcal{L}(\theta;\Gamma_i), 
\end{equation} 
where $\mathcal{J}$ defines the total loss computed as the sum of the single-step loss $\mathcal{L}(\theta;\Gamma_i)$ and a discount factor $\lambda$, which controls the influence of predictions far in the future \cite{Hansen2022tdmpc}. $\Gamma \sim B$ defines a trajectory  sampled from the replay buffer \textit{B}, and the single-step loss $\mathcal{L}(\theta;\Gamma_i)$ is defined as \cite{Hansen2022tdmpc}: 

\begin{equation}
\label{eq:singlestep}
    \mathcal{L}(\theta;\Gamma_i) = c_1 l^r_i + c_2 l^v_i + c_3 l^c_i,
\end{equation} 
where $l^r_i, l^v_i, l^c_i$ are the reward loss, value loss and consistency loss, respectively and defined as in \cite{Hansen2022tdmpc}:

\begin{equation}
\label{eq:reward}
l^r_i = || R_\theta(z_i,a_i) - r_i ||_2^2.
\end{equation} 
\begin{equation}
\label{eq:value}
l^v_i = || Q_\theta(z_i,a_i) - (r_i + \gamma Q_{\theta^-}(z_{i+1},\pi_\theta(z_{i+1}))) ||_2^2. 
\end{equation} 
\begin{equation}
\label{eq:consistency}
l^c_i = || d_\theta(z_i,a_i) - h_{\theta^-}(s_{i+1}) ||_2^2.
\end{equation} 

The single-step loss is then used to jointly optimize the reward, value, and dynamics functions with coefficients $c_1,c_2$ and $c_3$ balancing the weight of the 3 terms \cite{Hansen2022tdmpc}. During training, prioritized experience replay \cite{DBLP:journals/corr/SchaulQAS15} is used to store all experiences $(s_t,a_t,r_{t},s_{t+1})$ of the agent in a replay buffer, so that they can be used to train the TOLD model \cite{Hansen2022tdmpc}. Using the collected experiences, the TOLD model is iteratively updated throughout training to improve the predictions made by the value, reward, dynamics and policy functions. Given a sample trajectory $\Gamma \sim B$ from the replay buffer \textit{B} of length \textit{H}, we start at observation $s_t$ and consider this as the starting point. At first, $s_t$ is encoded into latent state $z_t$ using $h_\theta$. Next, using the dynamics function $d_{\theta}$, an \textit{H} time steps are unrolled by predicting the next latent state given the previous latent state and action. At each time step, the TOLD model predicts $(\hat{r}_t, \hat{q}_t, \hat{a}_t)$, and the single-step loss $\mathcal{L}(\theta;\Gamma_t)$ is calculated using Equation~\ref{eq:singlestep}. At the end, the total loss $\mathcal{J}$ is computed by summing the single-step losses over the \textit{H} time steps as indicated by Equation~\ref{eq:totalloss}, which is then used to perform one update of the model parameters. Then, the policy $\pi_\theta$ is updated by minimizing the objective 
\begin{equation}
\label{eq:policyloss}
\mathcal{J}_\pi(\theta;\Gamma_i) = -\sum_{i=t}^{t+H} \lambda^{i-t}Q_{\theta}(z_i,\pi_\theta(z_i)),
\end{equation} 
which is a time-weighted summation of the policy objective widely used in actor-critic methods \cite{pmlr-v80-haarnoja18b,journals/corr/LillicrapHPHETS15}.

\subsection{Planning}
 During planning, the TD-MPC framework makes use of a slightly adopted version of Model Predictive Path Integral (MPPI)\cite{doi:10.2514/1.G001921} control, which is a sampling-based model predictive control algorithm. We introduce $(\mu^0,\sigma^0)_{t:t+H}$, $\mu^0,\sigma^0 \in \mathbb{R}^m, A \in \mathbb{R}^m$, which defines parameters for a normal distribution $\mathcal{N}$ used to sample trajectories for planning \cite{Hansen2022tdmpc}. Besides the trajectories sampled from the normal distribution, a small number of trajectories are included generated by the learned policy $\pi_\theta$. Using the dynamics function $d_\theta$, all the sampled trajectories are unrolled and evaluated by estimating the total return as follows: 

\begin{equation}
G_\Gamma = \mathbb{E} \left[\gamma^H Q_\theta(z_H, a_H) + \sum_{t=0}^{H-1}\gamma^t R_\theta(z_t, a_t) \right].
\end{equation} 


The terminal Q-function $Q_\theta(z_H, a_H)$ provides an estimation of the return beyond the planning horizon \textit{H} which supplements the short-term estimation provided by the reward function $R_\theta$. Selecting only the best \textit{k} trajectories $\Gamma^*$ based on their return, parameters $\mu^j$ and $\sigma^j$ at iteration \textit{j} are iteratively updated using the following estimates which are normalized in terms of the top-\textit{k} returns:

\begin{equation}
\label{eq:musigma}
\mu^j = \frac{\sum_{i=1}^k \Omega_i \Gamma_i^*}{\sum_{i=1}^k \Omega_i}, \sigma^i = \max \left( \sqrt{\frac{\sum_{i=1}^k \Omega_i(\Gamma_i^* - \mu^j)^2}{\sum_{i=1}^k \Omega_i}}, \epsilon \right),
\end{equation}
where $\Omega_i = \tau * (G_i - G^*)$ \cite{Hansen2022tdmpc}. Additionally, $G_i$ denotes the return of trajectory $\Gamma_i$, $G^*$ represents the maximum return, $\tau$ is a temperature parameter regulating the influence of the best trajectories and $\epsilon$ is a linearly decayed constant that enforces constant exploration \cite{Hansen2022tdmpc}. This process of sampling trajectories and afterwards updating parameters $\mu$ and $\sigma$ is repeated for a total of \textit{J} iterations during one time step \cite{Hansen2022tdmpc}. Subsequently, one trajectory $\Gamma$ is sampled from the final return-normalized distribution $\mathcal{N}$ and the first action $a_t$ is executed. Afterwards, this planning process is repeated at the next time step \textit{t}+1 with a 1-step shifted mean $\mu$ from the previous time step \cite{Hansen2022tdmpc}. This prevents the agent from starting at 0 every time and instead provides a good starting point which also includes the learnt knowledge from previous experiences \cite{Hansen2022tdmpc}. However, a fairly high variance is chosen to avoid local minima. 

\section{TD-MPC with SRL}
\label{approach}

\subsection{Reconstruction Function}
While in TD-MPC, the latent state representation is mainly learned and influenced by the consistency loss, we propose to add another loss term to the single-step loss $\mathcal{L}(\theta;\Gamma_i)$ which aims to provide the agent with a richer learning signal and thus should also improve the performance of the entire model and algorithm. We extend the TOLD model by another learnable component $h_\theta^{-1}$ (shown in Fig.~\ref{fig:newTOLD}) defined as:
\begin{align*}
	\textbf{Reconstruction function} &: \mathbf{\tilde{s}_t = h_{\theta}^{-1}(z_t)},
\end{align*} 
where $h_\theta^{-1}$ represents the inverse of the latent state representation function $h_\theta$. We denote the result of $h_\theta^{-1}(z_t)$ as $\tilde{s}_t$ which represents an estimate of the respective ground truth observation $s_t$ since in most cases the real observation $s_t$ cannot be fully reconstructed. The reason for this is that the task-oriented latent state representation $z_t$ will most likely be far less complex and thus it would be challenging, if not even impossible, to reconstruct the original state $s_t$ entirely with only being given the latent state. As denoted by parameters $\theta$, this additional function will be learnt together with the other components of the TOLD model. Hence, error gradients from all 4 terms $l^r,l^v,l^c$ and $l^{h^{-1}}$ are back-propagated through time to update each component. This also means that the other functions are forced to include information which is important for the reconstruction of the latent representation. Given a trajectory $\Gamma = (s_t,a_t,r_t,s_{t+1})_{t:t+H}$ sampled from our Replay Buffer \textit{B} and latent state representation $z_t = h_\theta(s_t)$, we supplement the loss $\mathcal{L}(\theta;\Gamma_i)$ (Equation~\ref{eq:singlestep}) with the following self-supervised reconstruction loss term: 
\begin{equation}
l^{h^{-1}}_i = || h_{\theta}^{-1}(z_i) - s_i ||_2^2.
\end{equation} 

Similar to the reward loss (see Equation~\ref{eq:reward}), the true observation $s_t$ from $\Gamma$ is being compared to the reconstructed observation $\tilde{s}_t = h_\theta^{-1}(z_t)$ by computing the mean squared error. Thus, the new single-step loss $\mathcal{L}(\theta;\Gamma_i)$, is defined as:

\begin{figure}[btp]
	 \centerline{\includegraphics[width=0.9\textwidth]{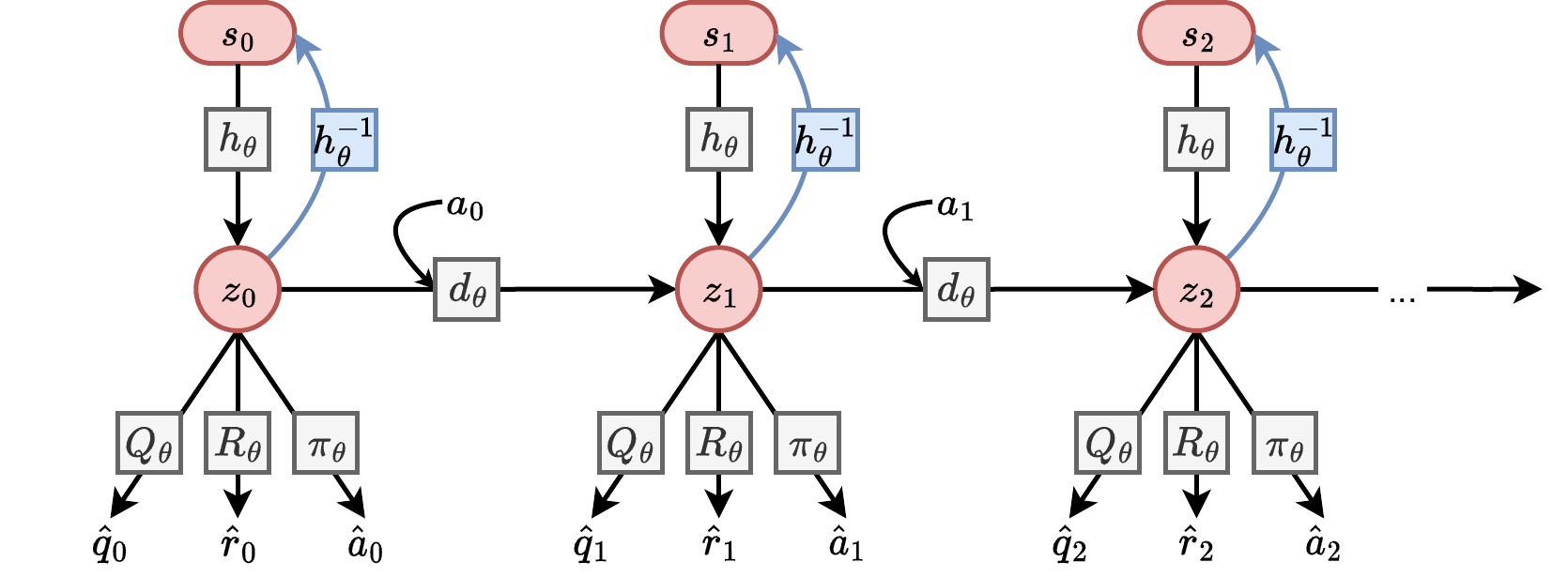}}
	 \caption{Updated TOLD model \cite{Hansen2022tdmpc}}
	 \label{fig:newTOLD}
\end{figure}
\begin{equation}
  \mathcal{L}(\theta;\Gamma_i) = c_1 l^r_i + c_2 l^v_i + c_3 l^c_i + c_4 l^{h^{-1}}_i,
\end{equation} 
where, $c_4$ is the reconstruction loss coefficient. As a result, the reconstruction function will not be used in any form within the planning process of the agent, but instead only utilized to stabilize the learning process. When considering a sparse reward signal, the learning process can be inherently challenging due to the reward serving as the primary learning signal in most cases. Thus, if rewards are sparse, so is the learning signal, and as a result, the agent potentially misses out on key insights about the environment dynamics or simply forgets about previously acquired knowledge. By including the term $l^{h^{-1}}$ in the single-step loss, we aim to further enhance the learning signal and thereby improving the robustness of the learning process, especially in sparse and noisy environments. Since the reconstruction function is also jointly trained with the other components, error gradients are  propagated through the other functions, resulting in shared behaviours and features. Thus, both the representation and dynamics functions will be forced to retain information essential to reconstruct the original observation. Including a reconstruction loss can potentially lead to the representation and dynamics functions gaining a better understanding of the environment beyond just the reward signal. Key features or other characteristics that help distinguish different states could be incorporated to further improve the learning process. The goal of the reconstruction loss is to augment the learning signal, without dominating the other loss terms.  

Another possible advantage of our proposed change to the TD-MPC framework is the ability to pre-train the world model in a self-supervised manner. Essentially, we can enable the agent to explore the environment without any reward or goal and use the reconstruction loss as the primary learning signal. As a result, the agent can acquire knowledge about the environments dynamics and construct a world model before the actual training process begins. This approach can be advantageous as having a pre-trained model of the environment allows for better estimates from the start and thus for reduced training times. 

\section{Experiments}

\subsection{Setup}
Similar to the experiments conducted in the TD-MPC paper, we run all environments with different random seeds and average the results \cite{Hansen2022tdmpc}. As the original TD-MPC has already been thoroughly compared to other state-of-the-art algorithms, we consider it sufficient to limit our comparison to the original TD-MPC framework. We use the documented results from the official GitHub repository\footnote{\url{https://github.com/nicklashansen/tdmpc}} for comparison of state-based environments, whereas we generate the data for image-based environments on our own using the official implementation, due to no data being available for most image-based tasks. We follow the same approach as in \cite{Hansen2022tdmpc} and conduct experiments on various environments from the DeepMind Control suite (DMControl)\cite{DBLP:journals/corr/abs-1801-00690}. Hyperparameter values that are not explicitly stated are set to the values in \cite{Hansen2022tdmpc}.

We adopt the TD-MPC framework as described in Section~\ref{approach}. Similar to the other components of the TOLD model, we employ a deterministic MLP to learn the reconstruction function. For state-based tasks, the architecture of the MLP consists of two simple fully-connected layers together with an ELU activation function, chosen intentionally to resemble the encoder architecture of the latent state representation function. The MLP architecture for image-based tasks is slightly more intricate due to the need of upsampling the given latent representation to match the original input size of 84x84 pixels. The initial layer is a Linear Layer, used to preprocess the input for the subsequent, whereas the next four hidden layers are deconvolutional layers, with all except the last being followed by a batch normalization layer to enhance the stability of the learning process. In terms of activation functions, we chose the ReLU function for all layers, except for the final layer, where we used a sigmoid function to ensure that the outputs fall within the range of $[0,1]$. Additional to normalizing the images, a pixel shift augmentation of $\pm4$ is used to prevent overfitting of the TOLD encoder $h_\theta$ \cite{Hansen2022tdmpc}. For the training of our introduced reconstruction function, we use the original images instead of the augmented ones within the respective loss $l^{h^{-1}}$ because we aim to reconstruct the original observation as provided by the environment. We empirically choose a reconstruction coefficient of 2.0, 0.25, 0.275 for the state-based Finger Turn Hard, Cheetah Run, and Acrobat-Swingup, and of 0.15, 0.275, 0.15 for the image-based Finger Turn Hard, Cheetah Run, and Reacher Easy environments respectively.

\subsection{Results}
To evaluate the performance of the two algorithms we conduct a periodic performance evaluation during training after every 20k environment steps. We evaluate the agent after every 20k environment steps, averaging over 10 episodes, then the averaged return is calculated. In the description of the results and conclusion, the term TD-MPC agent is used to refer to the agent using the original TD-MPC framework, while the TD-MPC framework with the additional reconstruction function is referred to as the reconstruction agent. Additionally, we measure the performance for image-based tasks after 100k and 500k environment steps to compare the initial learning speed, asymptotic performance as well as sample efficiency and robustness of the algorithms.

\begin{figure}[h]
     \centering
     \begin{subfigure}[b]{1\textwidth}
         \centering
         \includegraphics[width=\textwidth]{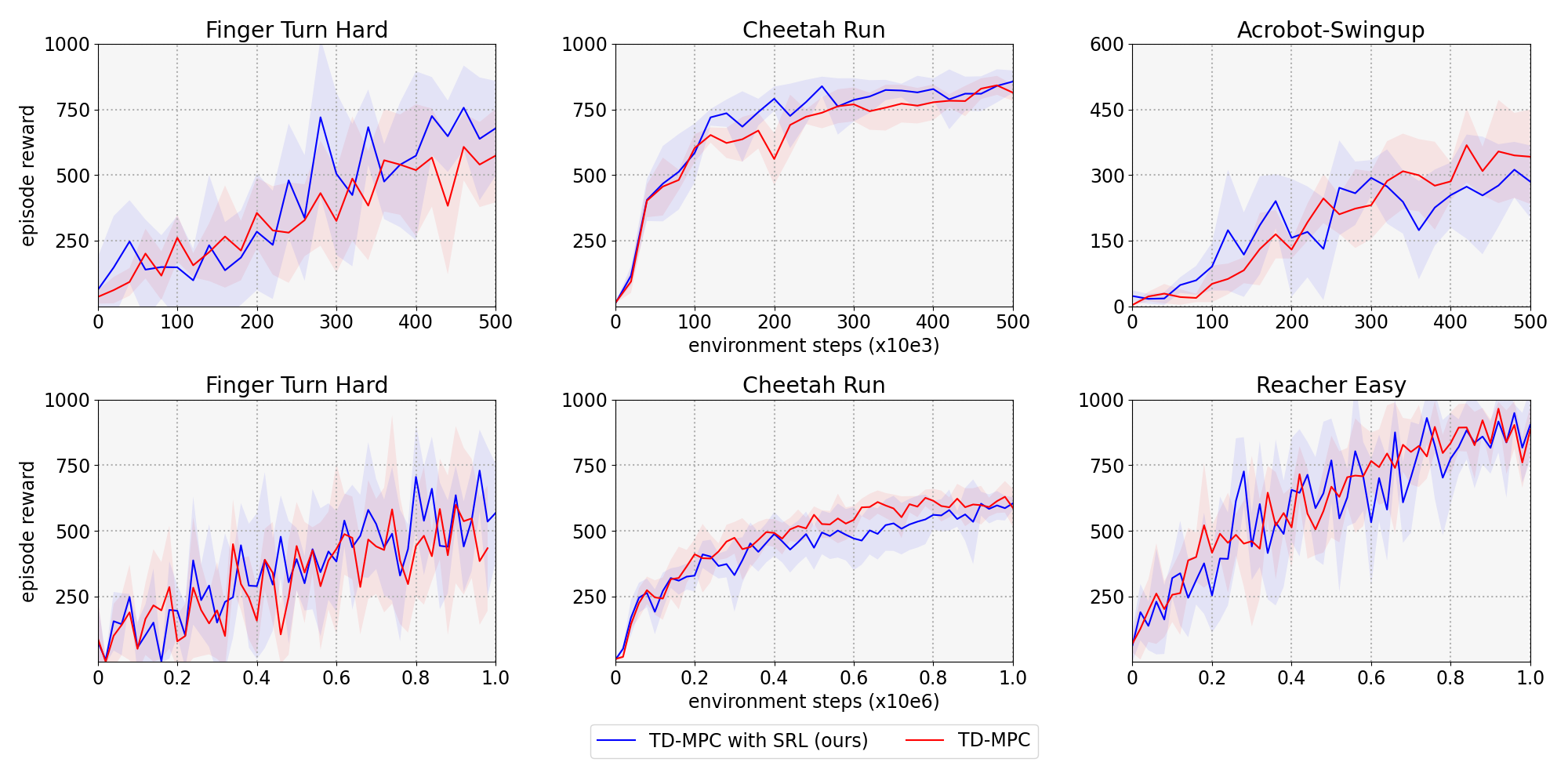}
     \end{subfigure}
        \caption{Episode return comparison of our method (TD-MPC with SRL) and the original TD-MPC framework on state-based (top row) and image-based (bottom row) environments. We follow the approach from the original TD-MPC paper \cite{Hansen2022tdmpc} and average the return over 5 runs.}
        \label{fig:performances}
\end{figure}

\begin{figure}[h]
     \centering
     \begin{subfigure}[b]{0.9\textwidth}
         \centering
         \includegraphics[width=\textwidth]{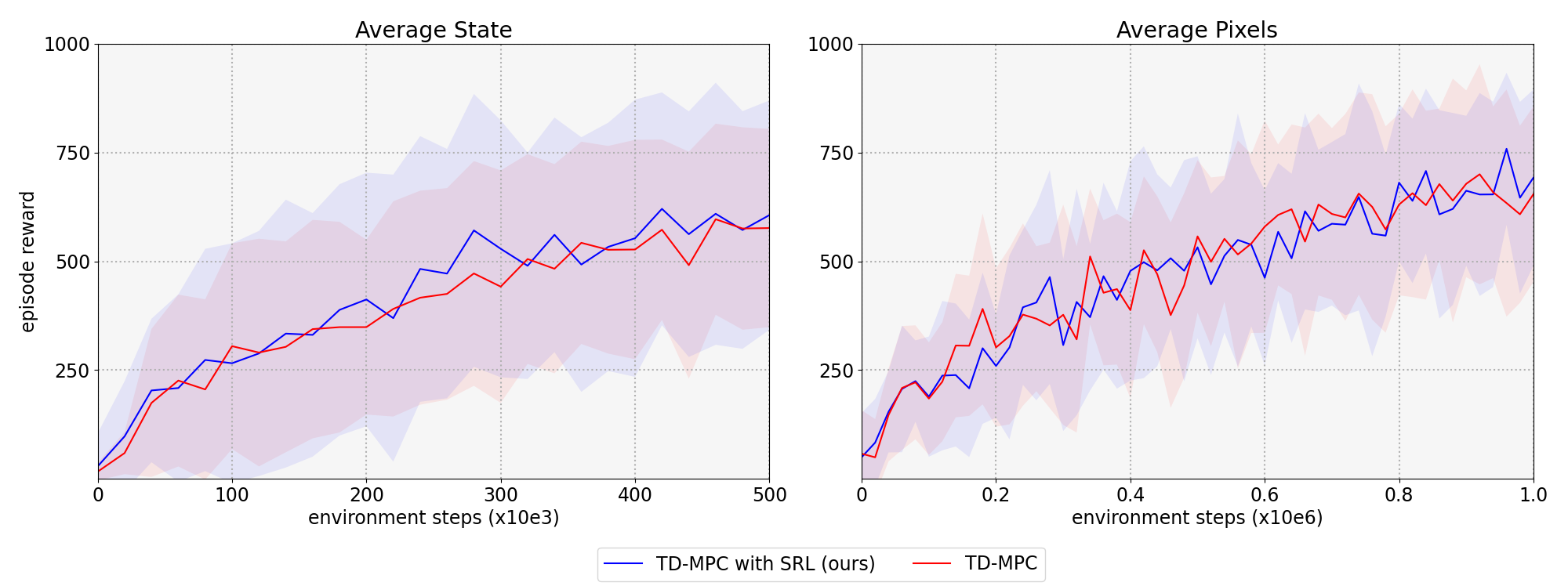}
     \end{subfigure}
    \caption{Episode return comparison of our method (TD-MPC with SRL) and the original TD-MPC framework. The left plot represents the average return over all three state-based environments, while the right does the same for the three image-based environments from Fig.~\ref{fig:performances}.}
    \label{fig:avg-perf}
\end{figure}

For state-based environments, the reconstruction agent outperforms the regular TD-MPC agent on two of the three selected tasks (see Fig.~\ref{fig:performances}). For the tasks Finger Turn Hard and Cheetah Run, the additional reconstruction function yields an increased learning speed and improved overall performance. In the Acrobot Swingup environment the reconstruction agent achieves a better initial learning speed, but worse asymptotic performance, while in terms of stability, we achieve comparable results in all three environments. Looking at the overall performance averaged over all three tasks, our changes to the TD-MPC framework slightly increase the learning speed and performance but seem to be slightly less stable when compared to the TD-MPC agent (see Fig.~\ref{fig:avg-perf}).

In the selected image-based environments, the changes made to the single-step loss term and the additional reconstruction function do not appear to have a significant impact on the overall learning speed of our agent (see Fig.~\ref{fig:performances}). Based on the results in Table~\ref{tb:500k100k}, the reconstruction agent demonstrated improved performance and stability after 100k environment steps on two of the tasks, which suggests a faster initial learning process and enhanced stability, while achieving comparable performance on the others. After 500k steps, our agent outperformed the original framework in only one of the tested environments but showed a significant improvement in terms of stability (lower standard deviation) on two of the three evaluated tasks. In general, our framework achieved comparable episode returns in most image-based environments, as shown in Fig.~\ref{fig:avg-perf}, which displays the average reward of the three image-based environments: Finger Turn Hard, Cheetah Run, and Reacher Easy. Overall, however, the modified agent exhibited improved stability after 500k environment steps in two of the three evaluated environments (Reacher Easy, Finger Turn Hard), as indicated in Table~\ref{tb:500k100k}.

\begin{table}[h]

\begin{center}
\begin{tabular}{l c c} 
 \hline
 100K STEP SCORES & TD-MPC & TD-MPC with SRL \\ [0.5ex] 
 \hline\hline
 Cheetah Run & \textbf{246} $\pm$ \textbf{73} & 191 $\pm$ 93  \\ 
 \hline
 Reacher Easy & 256 $\pm$ 116 & \textbf{319} $\pm$ \textbf{82} \\
 \hline
 Finger Turn Hard &  50 $\pm$ 100 & \textbf{55.3} $\pm$ \textbf{108}  \\
 \hline
 \hline
\end{tabular}
\end{center}

\begin{center}
\begin{tabular}{l c c} 
 \hline
 500K STEP SCORES & TD-MPC & TD-MPC  with SRL \\ [0.5ex] 
 \hline\hline
 Cheetah Run & \textbf{561} $\pm$ \textbf{35} & 435 $\pm$ 79 \\ 
 \hline
 Reacher Easy & 668 $\pm$ 217 & \textbf{768} $\pm$ \textbf{178} \\
 \hline
 Finger Turn Hard & \textbf{442} $\pm$ \textbf{187} & 392 $\pm$ 132  \\
 \hline
 \hline
 
\end{tabular}
 \caption{Episode return comparison for image-based environments after 100k and 500k environment steps with the respective standard deviations. The column with bold numbers indicates better results in terms of return.}
 \label{tb:500k100k}
\end{center}
\end{table}

\subsection{Discussion}
The state-based environments, especially Finger Turn Hard and Cheetah Run, showed the most significant improvements in terms of overall performance. The introduction of the reconstruction function and its associated loss term appears to help the agent with learning a more accurate TOLD model at a faster pace than its original counterpart. Learning the reconstruction function together with the other components also results in more frequent updates to the other components like the value and dynamics functions which could possibly achieve an increased learning speed and stability. However, the modified agent did not show significant improvements in stability, but rather a slight decrease, especially in simple low-dimensional environments like Finger Turn Hard or Acrobot Swingup, where the learning process was less robust and instead more susceptible to random seeds. This is potentially due to the unnecessary additional information being encoded in the latent state. On the other hand, for more complex environments like Cheetah Run, we observed a significantly more robust and slightly faster learning process, suggesting that the information needed for reconstruction is useful in this domain and can help with strengthening the learning signal. 

Learning from pixels proves to be more challenging for our agent, which may be due to their inherent noisy and high-dimensional nature. In the context of noisy environments, the role of the latent state is crucial as it reduces the complexity of the observation by encoding only the task-relevant details, thereby filtering out irrelevant information. Therefore, our modifications to the TD-MPC algorithm allow for a more efficient and stable learning process by enhancing the learning signal and updating the model components more frequently. This is important since the agent may miss out on important information due to sparse rewards when only trained on the reward signal. The increase in performance and stability are most likely attributed to the reduced susceptibility to noise and random seeds, resulting in a more consistent and steady learning process. However, in more complex environments such as Cheetah Run, the overall performance return-wise may still be slightly worse, due to the lack of hyperparameter tuning. To further validate our findings, a more diverse comparison of performance would include a variety of other complex environments, such as the Humanoid or Dog tasks. 

One aspect that we have not yet explored is the ability of the modified TD-MPC with self-supervised representation learning to pre-train a model of the environment, which could supply the agent with better initial estimates throughout training. Therefore, this could potentially be another valuable opportunity to further improve the performance and stability of our modified agent. We leave this open for future work.

\section{Conclusion}
In this paper, we proposed a modification of the TD-MPC framework, specifically the TOLD model. This included an additional self-supervised loss term, as well as a new component aimed at reconstructing the original observation given a latent state and thus providing the agent with an enhanced learning signal. We assessed the modified agent's performance in various experiments from the DMControl suite and compared it to the original version. Our findings showed that the modified agent performed better on the majority of state-based environments while achieving comparable stability. In image-based tasks, which inherently have high noise levels, the reconstruction agent proved to have a faster initial learning speed in most experiments, and despite struggling with the overall learning speed in most environments, the modified agent demonstrated a more robust learning process in the majority of tasks. Especially for high-dimensional environments, the agent is still struggling in terms of episode return and stability. Nevertheless, most importantly, we do not observe a significant decrease in terms of performance on any of the tasks. These results indicate the potential benefits of the reconstruction function and its associated loss term in generating an enhanced learning signal and consistent updating of the TOLD model components that is not dependent on reward.

\section{Acknowledgements}
The authors gratefully acknowledge support from the
DFG (CML, MoReSpace, LeCAREbot), BMWK (SIDIMO, VERIKAS), and the European
Commission (TRAIL, TERAIS). Mostafa Kotb is funded by a scholarship from the Ministry of Higher Education of the Arab Republic of Egypt.

%
%
%
\bibliographystyle{splncs04}
\bibliography{bibliography}

\end{document}